\documentclass[11pt,a4paper]{article}
\usepackage[hyperref]{acl2019}
\usepackage{times}
\usepackage{latexsym}
\usepackage{tabularx}
\usepackage{graphicx}
\usepackage{float}
\usepackage{enumerate}
\usepackage{amssymb,amsmath}
\usepackage{color}
\usepackage{bm}
\usepackage{makecell}
\usepackage{multi row}
\usepackage{array}
\usepackage{url}
\usepackage{resizegather}
\usepackage{refcount}
\usepackage{fixfoot}
\usepackage{placeins}

\newcolumntype{Y}{>{\centering\arraybackslash}X}
\newcommand\tstrut{\rule{0pt}{2.6ex}}         
\newcommand\bstrut{\rule[-1.0ex]{0pt}{0pt}}   
\newcommand{\thinline}{\Xhline{1.5\arrayrulewidth}}
\newcommand{\thickline}{\Xhline{2.5\arrayrulewidth}}
\newcommand{\tsep}	{\bstrut \\ \thinline}
\newcommand{\ttop}{\thickline}
\newcommand{\tbottom}{\bstrut \\ \thickline}

\newcommand{\xhdr}[1]{\vspace{1.7mm}\noindent{{\bf #1.}}}

\newcommand{\alns}[1] {
	\begin{align*} #1 \end{align*}
}
\newcommand{\head}{\textsc{H}}

\aclfinalcopy 

\title{What Does BERT Look At?\\An Analysis of BERT's Attention}
 \author{Kevin Clark$^\dagger$ \hspace{4mm} Urvashi Khandelwal$^\dagger$ \hspace{4mm} Omer Levy$^\ddagger$ \hspace{4mm} Christopher D. Manning$^\dagger$\\
  $^\dagger$Computer Science Department, Stanford University\\ $^\ddagger$Facebook AI Research \\
  {\tt \{kevclark,urvashik,manning\}@cs.stanford.edu} \\ {\tt omerlevy@fb.com}
}
 
\date{}

\begin{document}
\maketitle
\begin{abstract}
Large pre-trained neural networks such as BERT have had great recent success in NLP, motivating a growing body of research investigating what aspects of language they are able to learn from unlabeled data.
Most recent analysis has focused on model outputs (e.g., language model surprisal) or internal vector representations (e.g., probing classifiers). 
Complementary to these works, we propose methods for analyzing the attention mechanisms of pre-trained models and apply them to BERT. 
BERT's attention heads exhibit patterns such as attending to delimiter tokens, specific positional offsets, or broadly attending over the whole sentence, with heads in the same layer often exhibiting similar behaviors. 
We further show that certain attention heads correspond well to linguistic notions of syntax and coreference.
For example, we find heads that attend to the direct objects of verbs, determiners of nouns, objects of prepositions, and coreferent mentions with remarkably high accuracy.
Lastly, we propose an attention-based probing classifier and use it to further demonstrate that substantial syntactic information is captured in BERT's attention. 
\end{abstract}

\section{Introduction}
Large pre-trained language models achieve very high accuracy when fine-tuned on supervised tasks \citep{dai2015semi,peters2018deep,radford2018improving}, but it is not fully understood why.
The strong results suggest pre-training teaches the models about the structure of language, but what specific linguistic features do they learn?

Recent work has investigated this question by examining the {\it outputs} of language models on carefully chosen input sentences \citep{linzen2016assessing} or examining the internal {\it vector representations} of the model through methods such as probing classifiers \cite{Adi2017FinegrainedAO,Belinkov2017WhatDN}. 
Complementary to these approaches, we study\footnote{Code will be released at \url{https://github.com/clarkkev/attention-analysis}.} the {\it attention maps} of a pre-trained model. 
Attention \citep{bahdanau2014neural} has been a highly successful neural network component.
It is naturally interpretable because an attention weight has a clear meaning: how much a particular word will be weighted when computing the next representation for the current word.
Our analysis focuses on the 144 attention heads in BERT\footnote{We use the English base-sized model.} \citep{devlin2018bert}, a large pre-trained  Transformer \citep{Vaswani2017AttentionIA} network that has demonstrated excellent performance on many tasks. 

We first explore generally how BERT's attention heads behave.
We find that there are common patterns in their behavior, such as attending to fixed positional offsets or attending broadly over the whole sentence. 
A surprisingly large amount of BERT's attention focuses on the deliminator token [SEP], which we argue is used by the model as a sort of no-op.
Generally we find that attention heads in the same layer tend to behave similarly. 

We next probe each attention head for linguistic phenomena.
In particular, we treat each head as a simple no-training-required classifier that, given a word as input, outputs the most-attended-to other word.
We then evaluate the ability of the heads to classify various syntactic relations.
While no single head performs well at many relations, we find that particular heads correspond remarkably well to particular relations.
For example, we find heads that find direct objects of verbs, determiners of nouns, objects of prepositions, and objects of possessive pronouns with $>$75\% accuracy. 
We perform a similar analysis for coreference resolution, also finding a BERT head that performs quite well. 
These results are intriguing because the behavior of the attention heads emerges purely from self-supervised training on unlabeled data, without explicit supervision for syntax or coreference.

\begin{figure*}[!tb]
\begin{center}
\includegraphics[width=0.95\textwidth]{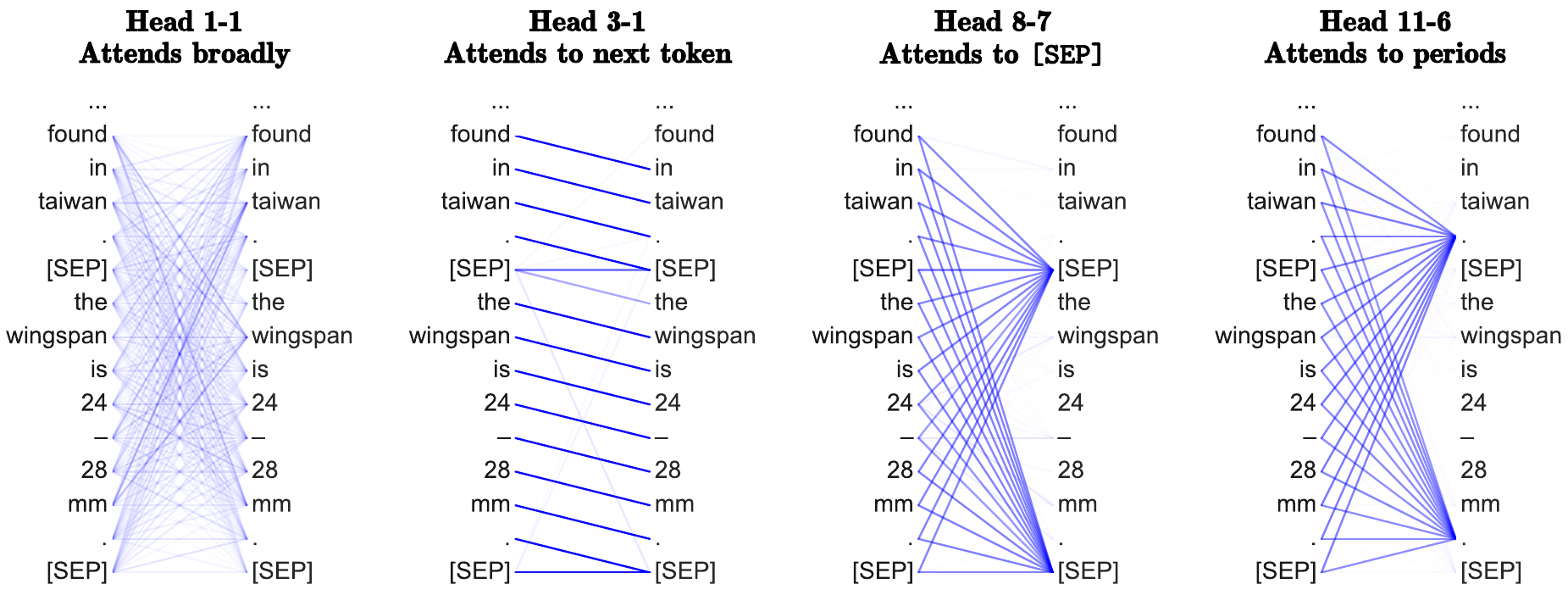}
\end{center}
\caption[cap]{Examples of heads exhibiting the patterns discussed in Section~\ref{sec:patterns}. The darkness of a line indicates the strength of the attention weight (some attention weights are so low they are invisible). }
\label{fig:patterns}
\end{figure*}

Our findings show that particular heads specialize to specific aspects of syntax.
To get a more overall measure of the attention heads' syntactic ability, we propose an attention-based probing classifier that takes attention maps as input.
The classifier achieves 77 UAS at dependency parsing, showing BERT's attention captures a substantial amount about syntax.
Several recent works have proposed incorporating syntactic information to improve attention \citep{Eriguchi2016TreetoSequenceAN, Chen2018SyntaxDirectedAF, strubell2018linguistically}.
Our work suggests that to an extent this kind of syntax-aware attention already exists in BERT, which may be one of the reason for its success.

\section{Background: Transformers and BERT}
Although our analysis methods are applicable to any model that uses an attention mechanism, in this paper we analyze BERT \citep{devlin2018bert}, a large Transformer \citep{Vaswani2017AttentionIA} network.
Transformers consist of multiple layers where each layer contains multiple attention heads.
An attention head takes as input a sequence of vectors $h = [h_1, ..., h_n]$ corresponding to the $n$ tokens of the input sentence.
Each vector $h_i$ is transformed into query, key, and value vectors $q_i, k_i, v_i$ through separate linear transformations. 
The head computes attention weights $\alpha$ between all pairs of words as softmax-normalized dot products between the query and key vectors. 
The output $o$ of the attention head is a weighted sum of the value vectors. 
\alns{
\alpha_{ij} = \frac{\exp{(q_i^T k_j)}}{\sum_{l=1}^n \exp{(q_i^T k_l)}} \quad\phantom{aa} o_i = \sum_{j=1}^n \alpha_{ij} v_j
}
Attention weights can be viewed as governing how ``important" every other token is when producing the next representation for the current token.  

BERT is pre-trained on 3.3 billion tokens of English text to perform two tasks.
In the ``masked language modeling" task, the model predicts the identities of words that have been masked-out of the input text.
In the ``next sentence prediction" task, the model predicts whether the second half of the input follows the first half of the input in the corpus, or is a random paragraph.
Further training the model on supervised data results in impressive performance across a variety of tasks ranging from sentiment analysis to question answering. 
An important detail of BERT is the preprocessing used for the input text. A special token [CLS] is added to the beginning of the text and another token [SEP] is added to the end. If the input consists of multiple separate texts (e.g., a reading comprehension example consists of a separate question and context), [SEP] tokens are also used to separate them. As we show in the next section, these special tokens play an important role in BERT's attention. 
We use the ``base" sized BERT model, which has 12 layers containing 12 attention heads each.
We use $<$layer$>$-$<$head\_number$>$ to denote a particular attention head.

\section{Surface-Level Patterns in Attention}
\label{sec:patterns}
Before looking at specific linguistic phenomena, we first perform an analysis of surface-level patterns in how BERT's attention heads behave. Examples of heads exhibiting these patterns are shown in Figure~\ref{fig:patterns}.

\xhdr{Setup} We extract the attention maps from BERT-base over 1000 random Wikipedia segments. We follow the setup used for pre-training BERT where each segment consists of at most 128 tokens corresponding to two consecutive paragraphs of Wikipedia (although we do not mask out input words or as in BERT's training). The input presented to the model is [CLS]$<$paragraph-1$>$[SEP]$<$paragraph-2$>$[SEP]. 

\begin{figure}[tb]
\begin{center}
\includegraphics[width=0.99\columnwidth]{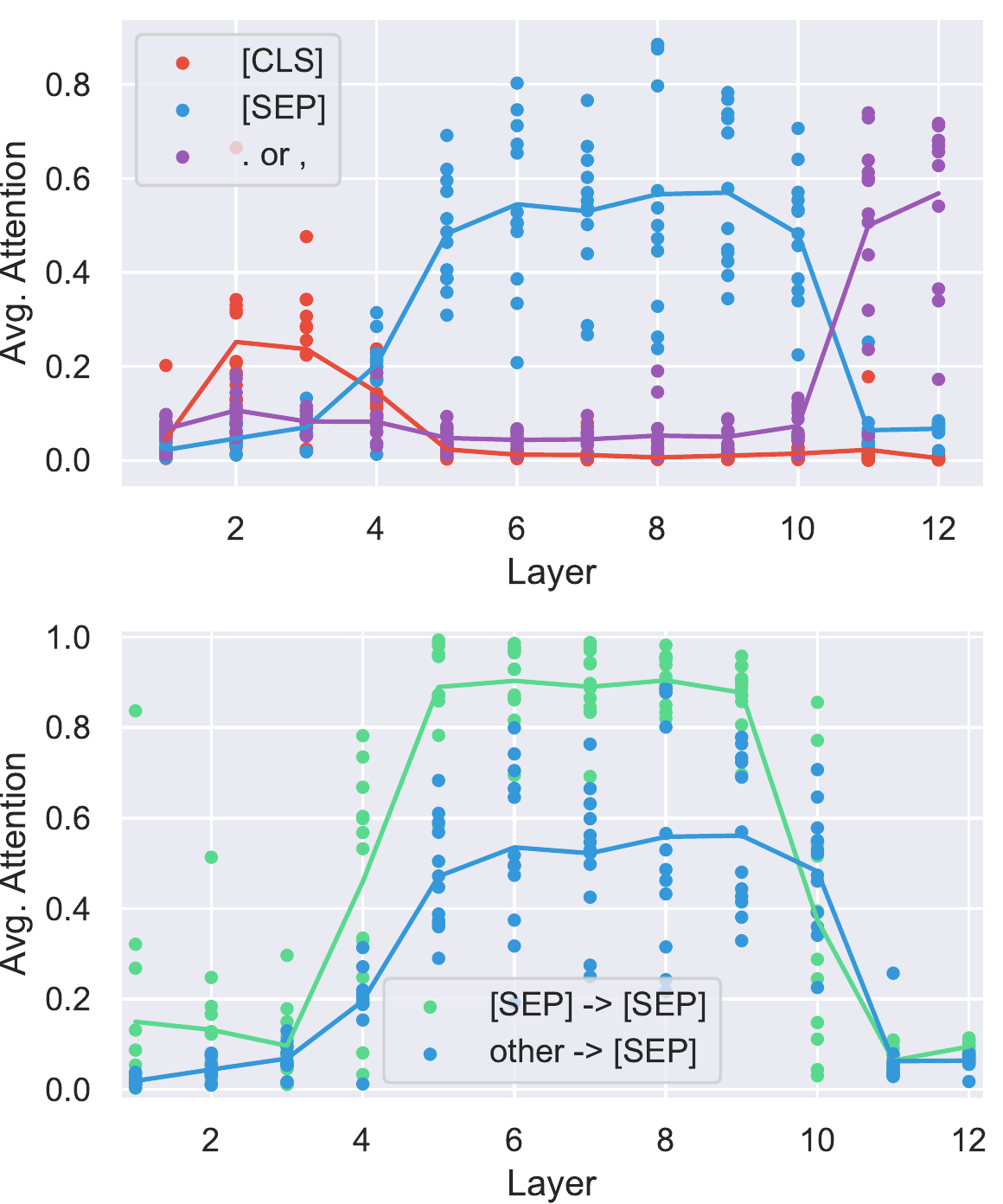}
\end{center}
\caption{Each point corresponds to the average attention a particular BERT attention head puts toward a token type. Above: heads often attend to ``special" tokens. Early heads attend to [CLS], middle heads attend to [SEP], and deep heads attend to periods and commas. Often more than half of a head's total attention is to these tokens. Below: heads attend to [SEP] tokens even more when the current token is [SEP] itself.}
\label{fig:tokens}
\end{figure}

\begin{figure}[t!]
\begin{center}
\includegraphics[width=0.99\columnwidth]{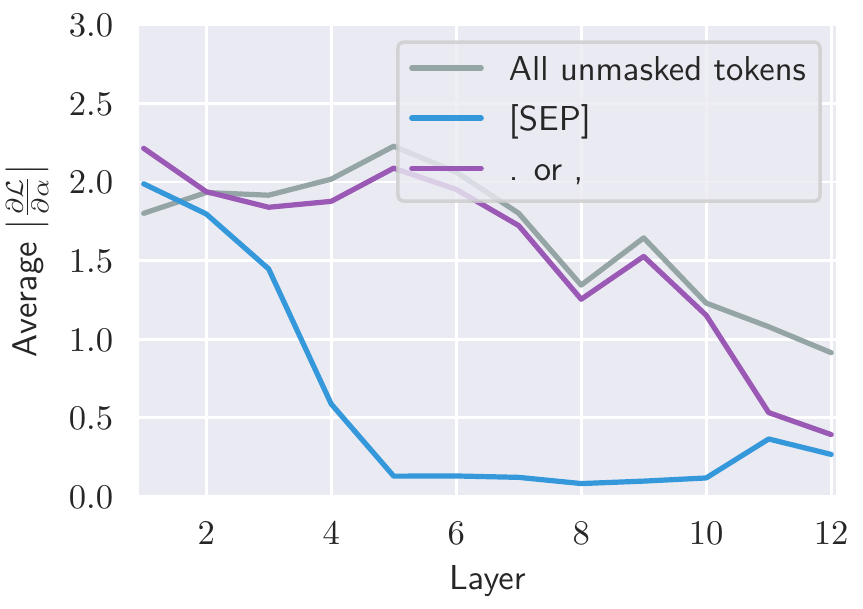}
\end{center}
\caption{Gradient-based feature importance estimates for attention to [SEP], periods/commas, and other tokens.  }
\label{fig:grad}
\end{figure}

\begin{figure}[t!]
\begin{center}
\includegraphics[width=0.99\columnwidth]{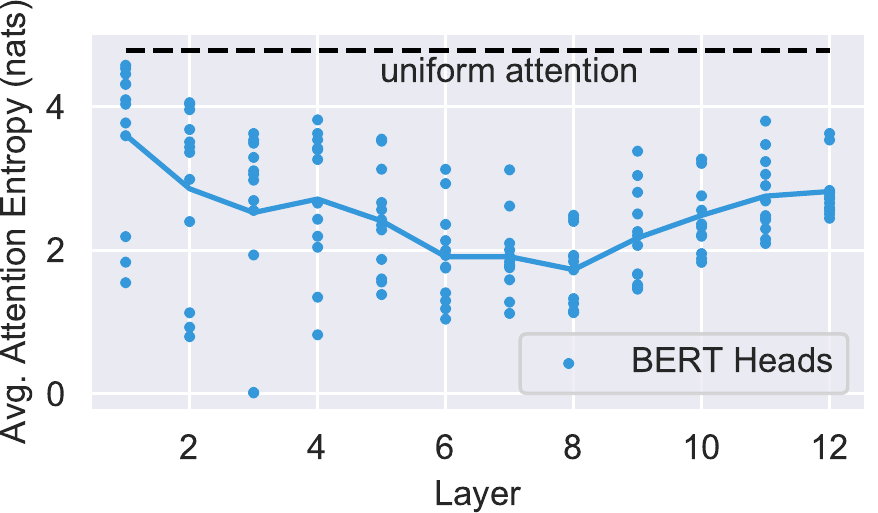}
\end{center}
\caption{Entropies of attention distributions. In the first layer there are particularly high-entropy heads that produce bag-of-vector-like representations. }
\label{fig:entropies}
\end{figure}

\subsection{Relative Position}
First, we compute how often BERT's attention heads attend to the current token, the previous token, or the next token.
We find that most heads put little attention on the current token.
However, there are heads that specialize to attending heavily on the next or previous token, especially in earlier layers of the network.
In particular four attention heads (in layers 2, 4, 7, and 8) on average put $>$50\% of their attention on the previous token and five attention heads (in layers 1, 2, 2, 3, and 6) put $>$50\% of their attention on the next token.

\subsection{Attending to Separator Tokens}
Interestingly, we found that a substantial amount of BERT's attention focuses on a few tokens (see Figure~\ref{fig:tokens}). 
For example, over half of BERT's attention in layers 6-10 focuses on [SEP].
To put this in context, since most of our segments are 128 tokens long, the average attention for a token occurring twice in a segments like [SEP] would normally be around 1/64.
[SEP] and [CLS] are guaranteed to be present and are never masked out, while periods and commas are the most common tokens in the data excluding ``the," which might be why the model treats these tokens differently.
A similar pattern occurs for the uncased BERT model, suggesting there is a systematic reason for the attention to special tokens rather than it being an artifact of stochastic training.

One possible explanation is that [SEP] is used to aggregate segment-level information which can then be read by other heads. 
However, further analysis makes us doubtful this is the case.
If this explanation were true, we would expect attention heads processing [SEP] to attend broadly over the whole segment to build up these representations. 
However, they instead almost entirely (more than 90\%; see bottom of Figure~\ref{fig:tokens}) attend to themselves and the other [SEP] token.
Furthermore, qualitative analysis (see Figure~\ref{fig:examples}) shows that heads with specific functions attend to [SEP] when the function is not called for.
For example, in head 8-10 direct objects attend to their verbs. For this head, non-nouns mostly attend to [SEP].
Therefore, we speculate that attention over these special tokens might be used as a sort of ``no-op" when the attention head's function is not applicable.

To further investigate this hypothesis, we apply gradient-based measures of feature importance \citep{sundararajan2017axiomatic}. In particular, we compute the magnitude of the gradient of the loss from BERT's masked language modeling task with respect to each attention weight. Intuitively, this value measures how much changing the attention to a token will change BERT's outputs. 
Results are shown in Figure~\ref{fig:grad}. Starting in layer 5 -- the same layer where attention to [SEP] becomes high -- the gradients for attention to [SEP] become very small. This indicates that attending more or less to [SEP] does not substantially change BERT's outputs, supporting the theory that attention to [SEP] is used as a no-op for attention heads.

\subsection{Focused vs Broad Attention}
Lastly, we measure whether attention heads focus on a few words or attend broadly over many words. 
To do this, we compute the average entropy of each head's attention distribution (see Figure~\ref{fig:entropies}).
We find that some attention heads, especially in lower layers, have very broad attention.
These high-entropy attention heads typically spend at most 10\% of their attention mass on any single word. 
The output of these heads is roughly a bag-of-vectors representation of the sentence. 

We also measured entropies for all attention heads from only the [CLS] token.
While the average entropies from [CLS] for most layers are very close to the ones shown in Figure~\ref{fig:entropies}, the last layer has a high entropy from [CLS] of 3.89 nats, indicating very broad attention. 
This finding makes sense given that the representation for the [CLS] token is used as input for the ``next sentence prediction" task during pre-training, so it attends broadly to aggregate a representation for the whole input in the last layer.

\section{Probing Individual Attention Heads}

Next, we investigate individual attention heads to probe what aspects of language they have learned.
In particular, we evaluate attention heads on labeled datasets for tasks like dependency parsing.
An overview of our results is shown in Figure~\ref{fig:examples}.

\subsection{Method}

We wish to evaluate attention heads at word-level tasks, but BERT uses byte-pair tokenization \citep{Sennrich2016NeuralMT}, which means some words ($\sim$8\% in our data) are split up into multiple tokens.
We therefore convert token-token attention maps to word-word attention maps.
For attention {\it to} a split-up word, we sum up the attention weights over its tokens. 
For attention {\it from} a split-up word, we take the mean of the attention weights over its tokens. 
These transformations preserve the property that the attention from each word sums to 1.
For a given attention head and word, we take whichever other word receives the most attention weight as that model's prediction\footnote{We ignore [SEP] and [CLS], although in practice this does not significantly change the accuracies for most heads.}

\subsection{Dependency Syntax}
\textbf{Setup.}
We extract attention maps from BERT on the Wall Street Journal portion of the Penn Treebank \citep{marcus1993building} annotated with Stanford Dependencies.  
We evaluate both ``directions" of prediction for each attention head: the head word attending to the dependent and
the dependent attending to the head word. 
Some dependency relations are simpler to predict than others: for example a noun's determiner is often the immediately preceding word. 
Therefore as a point of comparison, we show predictions from a simple fixed-offset baseline.
For example, a fixed offset of -2 means the word two positions to the left of the dependent is always considered to be the head.

\begin{table}
\begin{tabularx}{\columnwidth}{X Y Y Y Y}
\ttop
\textbf{Relation} & \textbf{Head} & \textbf{Accuracy} & \textbf{Baseline} \tstrut \tsep
All & 7-6$\phantom{0}$ & 34.5 & 26.3 (1)\tstrut\\
\texttt{prep} & 7-4$\phantom{0}$ & 66.7 & 61.8 (-1)\\
\texttt{pobj} & 9-6$\phantom{0}$ & \textbf{76.3} & 34.6 (-2)\\
\texttt{det} & 8-11 & \textbf{94.3} & 51.7 (1)\\
\texttt{nn} & 4-10 & 70.4 & 70.2 (1)\\
\texttt{nsubj} & 8-2$\phantom{0}$ & 58.5 & 45.5 (1)\\
\texttt{amod} & 4-10 & 75.6 & 68.3 (1)\\
\texttt{dobj} & 8-10 & \textbf{86.8} & 40.0 (-2)\\
\texttt{advmod} & 7-6$\phantom{0}$ & 48.8 & 40.2 (1)\\
\texttt{aux} & 4-10 & 81.1 & 71.5 (1)  \tsep 
\texttt{poss} & 7-6$\phantom{0}$ & \textbf{80.5} & 47.7 (1) \tstrut \\
\texttt{auxpass} & 4-10$\phantom{0}$ & \textbf{82.5} & 40.5 (1) \\
\texttt{ccomp} & 8-1$\phantom{0}$ & \textbf{48.8} & 12.4 (-2)\\
\texttt{mark} & 8-2$\phantom{0}$ & \textbf{50.7} & 14.5 (2) \\
\texttt{prt} & 6-7$\phantom{0}$ & \textbf{99.1} & 91.4 (-1)
\tbottom
\end{tabularx}
\caption{
The best performing attentions heads of BERT on WSJ dependency parsing by dependency type.
Numbers after baseline accuracies show the best offset found (e.g., (1) means the word to the right is predicted as the head). We show the 10 most common relations as well as 5 other ones attention heads do well on. Bold highlights particularly effective heads.} 
\label{tab:relns}
\end{table}

\begin{figure*}[tbh!]
\begin{center}
\includegraphics[width=0.98\textwidth]{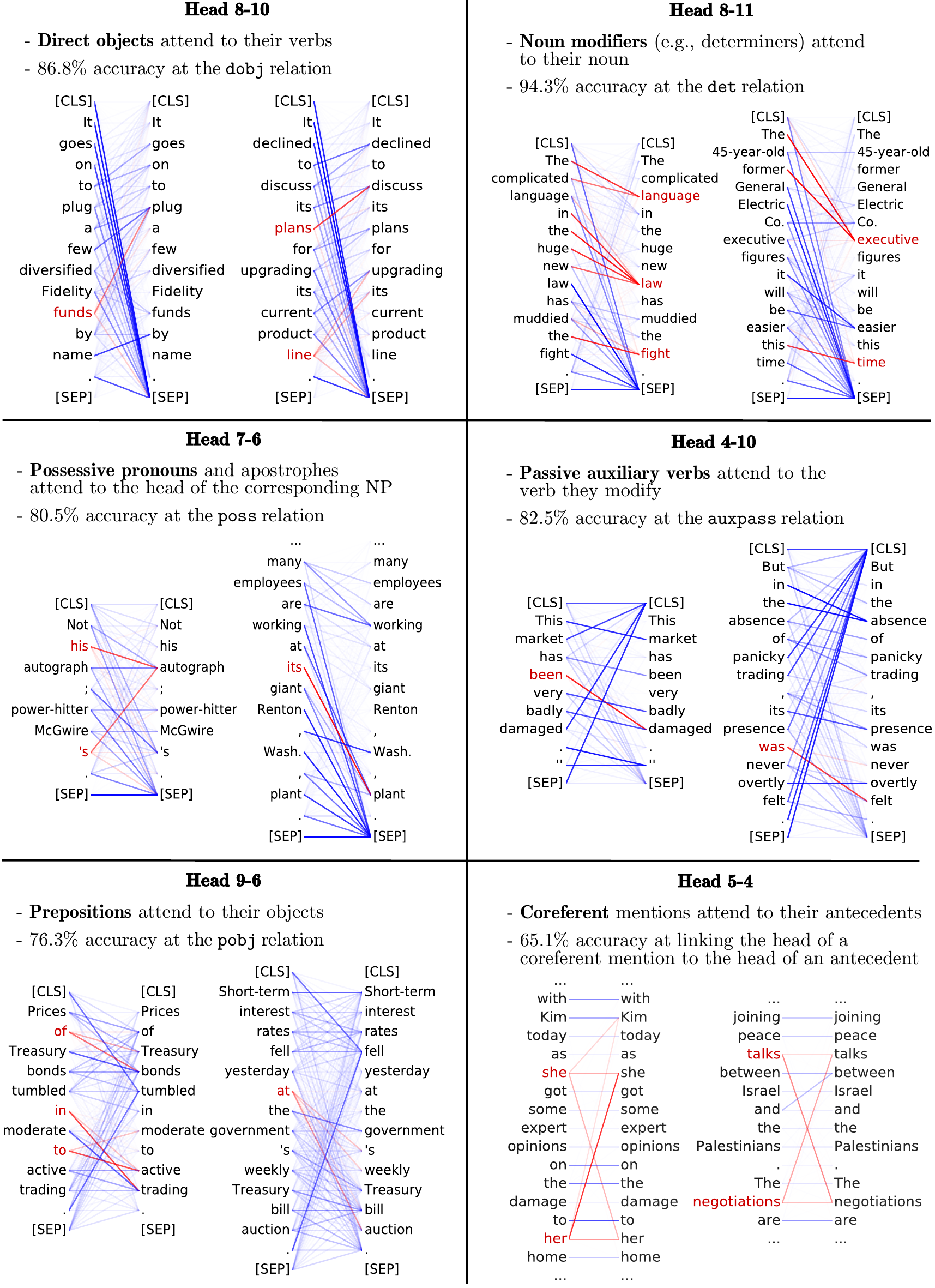}
\end{center}
\caption{BERT attention heads that correspond to linguistic phenomena. In the example attention maps, the darkness of a line indicates the strength of the attention weight. All attention to/from red words is colored red; these colors are there to highlight certain parts of the attention heads' behaviors. For Head 9-6, we don't show attention to [SEP] for clarity. Despite not being explicitly trained on these tasks, BERT's attention heads perform remarkably well, illustrating how syntax-sensitive behavior can emerge from self-supervised training alone. }
\label{fig:examples}
\end{figure*}

\xhdr{Results}
Table~\ref{tab:relns} shows that there is no single attention head that does well at syntax ``overall"; the best head gets 34.5 UAS, which is not much better than the right-branching baseline, which gets 26.3 UAS. This finding is similar to the one reported by \citet{Raganato2018AnAO}, who also evaluate individual attention heads for syntax.

However, we do find that certain attention heads specialize to specific dependency relations, sometimes achieving high accuracy and substantially outperforming the fixed-offset baseline. 
We find that for all relations in Table~\ref{tab:relns} except \texttt{pobj}, the dependent attends to the head word rather than the other way around, likely because each dependent has exactly one head but heads have multiple dependents.
We also note heads can disagree with standard annotation conventions while still performing syntactic behavior. 
For example, head 7-6 marks \emph{'s} as the dependent for the \texttt{poss} relation, while gold-standard labels mark the complement of an \emph{'s} as the dependent (the accuracy in Table~\ref{tab:relns} counts \emph{'s} as correct). 
Such disagreements highlight how these syntactic behaviors in BERT are learned as a by-product of self-supervised training, not by copying a human design.

Figure~\ref{fig:examples} shows some examples of the attention behavior. 
While the similarity between machine-learned attention weights and human-defined syntactic relations are striking, we note these are relations for which attention heads do particularly well on. 
There are many relations for which BERT only slightly improves over the simple baseline, so we would not say individual attention heads capture dependency structure as a whole. 
We think it would be interesting future work to extend our analysis to see if the relations well-captured by attention are similar or different for other languages.

\subsection{Coreference Resolution}

Having shown BERT attention heads reflect certain aspects of syntax, we now explore using attention heads for the more challenging semantic task of coreference resolution. Coreference links are usually longer than syntactic dependencies and state-of-the-art systems generally perform much worse at coreference compared to parsing. 

\xhdr{Setup}
We evaluate the attention heads on coreference resolution using the CoNLL-2012 dataset\footnote{We truncate documents to 128 tokens long to keep memory usage manageable.} \citep{pradhan2012conll}. In particular, we compute antecedent selection accuracy: what percent of the time does the head word of a coreferent mention most attend to the head of one of that mention's antecedents. We compare against three baselines for selecting an antecedent:
\begin{itemize}
    \itemsep0em
    \item Picking the nearest other mention.
    \item Picking the nearest other mention with the same head word as the current mention.
    \item A simple rule-based system inspired by \citet{lee2011stanford}. It proceeds through 4 sieves: (1) full string match, (2) head word match, (3) number/gender/person match, (4) all other mentions. The nearest mention satisfying the earliest sieve is returned.
\end{itemize}
We also show the performance of a recent neural coreference system from \citet{wiseman2015learning}.

\xhdr{Results}
Results are shown in Table~\ref{tab:coref}. We find that one of BERT's attention heads achieves decent coreference resolution performance, improving by over 10 accuracy points on the string-matching baseline and performing close to the rule-based system. It is particularly good with nominal mentions, perhaps because it is capable of fuzzy matching between synonyms as seen in the bottom right of Figure~\ref{fig:examples}.

\addtolength{\tabcolsep}{-4pt}
\begin{table}
\begin{tabularx}{\columnwidth}{X c c c c}
\ttop 
\textbf{Model} & \textbf{All} & \textbf{Pronoun} & \textbf{Proper} & \textbf{Nominal}\tstrut \tsep
Nearest & 27 & 29 & 29 & 19\tstrut \\
Head match & 52 & 47 & 67 & 40 \\
Rule-based & 69 & 70 & 77 & 60 \\
Neural coref & ~~83* & -- & -- & -- \tsep
Head 5-4 & 65 & 64 & 73 & 58\tstrut
\tbottom
\end{tabularx}
\vspace{1mm}\\
\small *Only roughly comparable because on non-truncated documents and with different mention detection.

\caption{Accuracies (\%) for systems at selecting a correct antecedent given a coreferent mention in the CoNLL-2012 data. One of BERT's attention heads performs fairly well at coreference.
}
\label{tab:coref}
\end{table}
\addtolength{\tabcolsep}{4pt}

\section{Probing Attention Head Combinations}

Since individual attention heads specialize to particular aspects of syntax, the model's overall ``knowledge" about syntax is distributed across multiple attention heads.
We now measure this overall ability by proposing a novel family of attention-based probing classifiers and applying them to dependency parsing.
For these classifiers we treat the BERT attention outputs as fixed, i.e., we do not back-propagate into BERT and only train a small number of parameters. 

The probing classifiers are basically graph-based dependency parsers. Given an input word, the classifier produces a probability distribution over other words in the sentence indicating how likely each other word is to be the syntactic head of the current one. 

\xhdr{Attention-Only Probe} Our first probe learns a simple linear combination of attention weights.
\alns{
    p(i|j) \propto \exp{\bigg(\sum_{k=1}^{n} w_k\alpha^k_{ij} + u_k \alpha^k_{ji}\bigg)}
}
where $p(i|j)$ is the probability of word $i$ being word $j$'s syntactic head, $\alpha^k_{ij}$ is the attention weight from word $i$ to word $j$ produced by head $k$, and $n$ is the number of attention heads. We include both directions of attention: candidate head to dependent as well as dependent to candidate head. The weight vectors $w$ and $u$ are trained using standard supervised learning on the train set. 

\xhdr{Attention-and-Words Probe} Given our finding that heads specialize to particular syntactic relations, we believe probing classifiers should benefit from having information about the input words. In particular, we build a model that sets the weights of the attention heads based on the GloVe \citep{pennington2014glove} embeddings for the input words. Intuitively, if the dependent and candidate head are ``the" and ``cat," the probing classifier should learn to assign most of the weight to the head 8-11, which achieves excellent performance at the determiner relation. The attention-and-words probing classifier assigns the probability of word $i$ being word $j$'s head as
\alns{
    p(i|j) \propto \exp \bigg( \sum_{k=1}^{n} &W_{k,:}(v_i \oplus v_j)\alpha^k_{ij} +\\  &U_{k,:}(v_i \oplus v_j) \alpha^k_{ji} \bigg)
}
Where $v$ denotes GloVe embeddings and $\oplus$ denotes concatenation. The GloVe embeddings are held fixed in training, so only the two weight matrices $W$ and $U$ are learned. The dot product $W_{k,:}(v_i \oplus v_j)$ produces a word-sensitive weight for the particular attention head. 

\xhdr{Results}
We evaluate our methods on the Penn Treebank dev set annotated with Stanford dependencies. We compare against three baselines:
\begin{itemize}
   \itemsep0em
    \item A right-branching baseline that always predicts the head is to the dependent's right.
    \item A simple one-hidden-layer network that takes as input the GloVe embeddings for the dependent and candidate head as well as distance features between the two words.\footnote{Indicator features for short distances as well as continuous distance features, with distance ahead/behind treated separately to capture word order}
    \item Our attention-and-words probe, but with attention maps from a BERT network with pre-trained word/positional embeddings but randomly initialized other weights. This kind of baseline is surprisingly strong at other probing tasks \citep{conneau2018you}.
\end{itemize}

Results are shown in Table~\ref{tab:probe}. 
We find the Attn + GloVe probing classifier substantially outperforms our baselines and achieves a decent UAS of 77, suggesting BERT's attention maps have a fairly thorough representation of English syntax. 

As a rough comparison, we also report results from the structural probe from \citet{hewitt2019structural}, which builds a probing classifier on top of BERT's vector representations rather than attention. 
The scores are not directly comparable because the structural probe only uses a single layer of BERT, produces undirected rather than directed parse trees, and is trained to produce the syntactic distance between words rather than directly predicting the tree structure. Nevertheless, the similarity in score to our Attn + Glove probing classifier suggests there is not much more syntactic information in BERT's vector representations compared to its attention maps.

Overall, our results from probing both individual and combinations of attention heads suggest that BERT learns some aspects syntax purely as a by-product of self-supervised training.
Other work has drawn a similar conclusions from examining BERT's predictions on agreement tasks \citep{goldberg2019assessing} or internal vector representations \citep{hewitt2019structural,Liu2019LinguisticKA}.
Traditionally, syntax-aware models have been developed through architecture design (e.g., recursive neural networks) or from direct supervision from human-curated treebanks. 
Our findings are part of a growing body of work indicating that indirect supervision from rich pre-training tasks like language modeling can also produce models sensitive to language's hierarchical structure.

\begin{table}
\begin{tabularx}{\columnwidth}{X >{\hsize=1.9cm}Y}
\ttop 
\textbf{Model} & \textbf{UAS} \tstrut \tsep
%\multirow{2}{*}{Structural probe \\\citep{hewitt2019structural}} & \multirow{2}{*}{80 UUAS*} \tsep
Structural probe & %\citep{hewitt2019structural} &
 80 UUAS*\tstrut \tsep
%\multirow{2}{*}{80 UUAS*} \\ \vspace{-5mm}\citep{hewitt2019structural} & \\
Right-branching & 26 \tstrut\\
Distances + GloVe & 58\\
Random Init Attn + GloVe & 30  \\
Attn  & 61 \\
Attn + GloVe & 77%\tsep
 %\multirow{2}{*}{80 UUAS*} \tstrut%\\ \citep{hewitt2019structural} & 
\tbottom
\end{tabularx}
\caption{Results of attention-based probing classifiers on dependency parsing. A simple model taking BERT attention maps and GloVe embeddings as input performs quite well. *Not directly comparable to our numbers; see text.}
\label{tab:probe}
\end{table}

\begin{figure}[tbh!]
\begin{center}
\includegraphics[width=0.99\columnwidth]{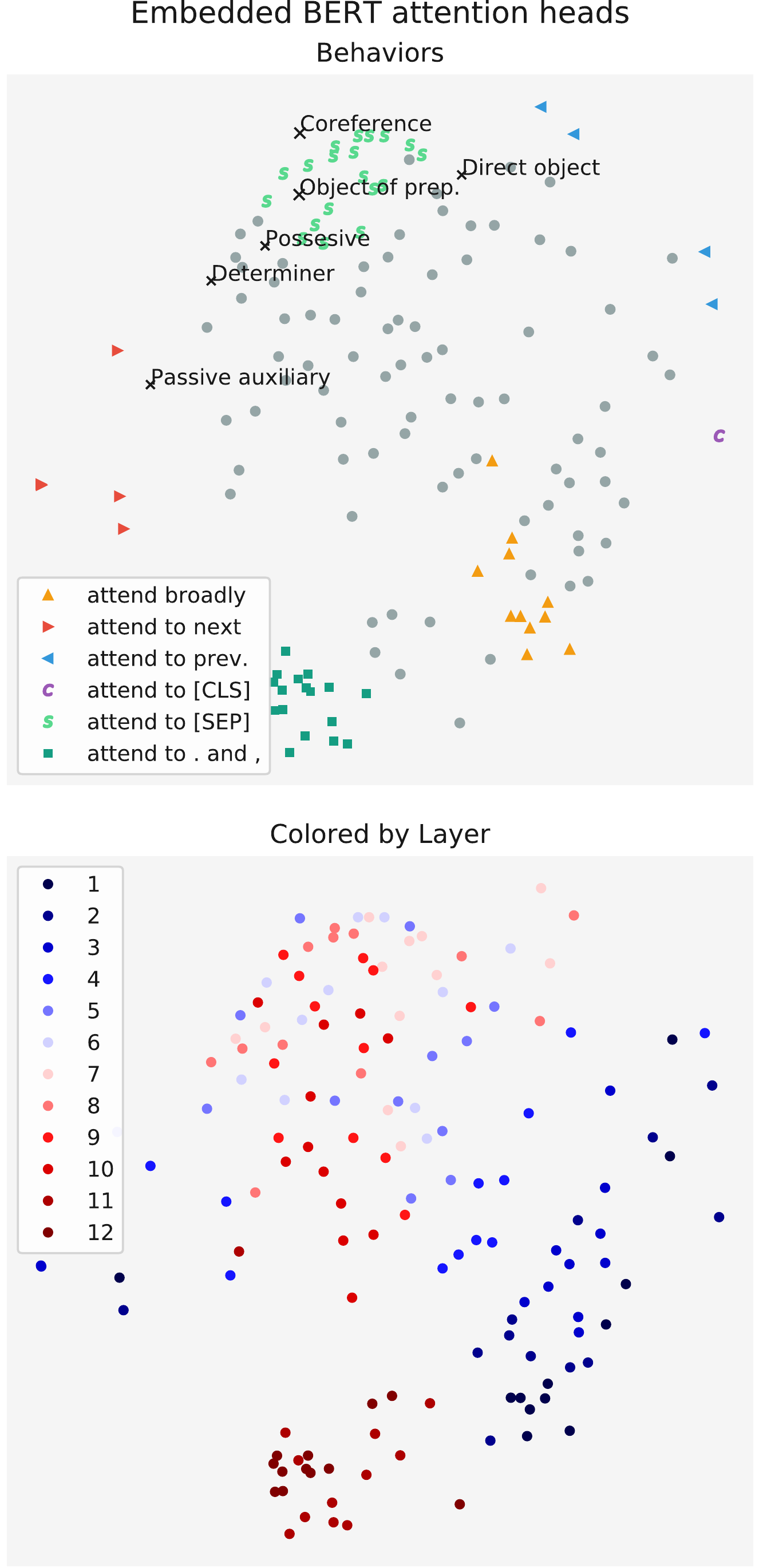}
\end{center}
\caption{BERT attention heads embedded in two-dimensional space. 
Distance between points approximately matches the average Jensen-Shannon divergences between the outputs of the corresponding heads. 
Heads in the same layer tend to be close together.
Attention head ``behavior" was found through the analysis methods discussed throughout this paper.}
\label{fig:embed}
\end{figure}

\section{Clustering Attention Heads}

Are attention heads in the same layer similar to each other or different? Can attention heads be clearly grouped by behavior? We investigate these questions by computing the distances between all pairs of attention heads. Formally, we measure the distance between two heads $\head_i$ and $\head_j$ as:
\alns{
    \sum_{\text{token} \in \text{data}} JS(\head_i(\text{token}), \head_j(\text{token}))
}
Where $JS$ is the Jensen-Shannon Divergence between attention distributions. 
Using these distances, we visualize the attention heads by applying multidimensional scaling \citep{kruskal1964multidimensional} to embed each  head in two dimensions such that the Euclidean distance between embeddings reflects the Jensen-Shannon distance between the corresponding heads as closely as possible. 

Results are shown in Figure~\ref{fig:embed}.
We find that there are several clear clusters of heads that behave similarly, often corresponding to behaviors we have already discussed in this paper.
Heads within the  same layer are often fairly close to each other, meaning that heads within the layer have similar attention distributions. 
This finding is a bit surprising given that \citet{Tu2018MultiHeadAW} show that encouraging attention heads to have different behaviors can improve Transformer performance at machine translation. 
One possibility for the apparent redundancy in BERT's attention heads is the use of attention dropout, which causes some attention weights to be zeroed-out during training.

\section{Related Work}

There has been substantial recent work performing analysis to better understand what neural networks learn, especially from language model pre-training. 
One line of research examines the {\it outputs} of language models on carefully chosen input sentences \citep{linzen2016assessing,Khandelwal2018SharpNF,Gulordava2018ColorlessGR,marvin2018targeted}.
For example, the model's performance at subject-verb agreement (generating the correct number of a verb far away from its subject) provides a measure of the model's syntactic ability, although it does not reveal {\it how} that ability is captured by the network.  

Another line of work investigates the internal {\it vector representations} of the model  \citep{Adi2017FinegrainedAO,giulianelli2018under,Zhang2018LanguageMT}, often using probing classifiers.
Probing classifiers are simple neural networks that take the vector representations of a pre-trained model as input and are trained to do a supervised task (e.g., part-of-speech tagging).
If a probing classifier achieves high accuracy, it suggests that the input representations reflect the corresponding aspect of language (e.g., low-level syntax).
Like our work, some of these studies have also demonstrated models capturing aspects of syntax \citep{Shi2016DoesSN,Blevins2018DeepRE} or coreference \citep{tenney2018you,tenney2019bert,Liu2019LinguisticKA} without explicitly being trained for the tasks. 

With regards to analyzing attention, \citet{vig2019visualizing} builds a visualization tool for the BERT's attention and reports observations about the attention behavior, but does not perform quantitative analysis. 
\citet{Burns2018ExploitingAT} analyze the attention of memory networks to understand model performance on a question answering dataset. %; we instead aim to understand linguistic information captured in pre-trained models.
There has also been some initial work in correlating attention with syntax. 
\citet{Raganato2018AnAO} evaluate the attention heads of a machine translation model on dependency parsing, but only report overall UAS scores instead of investigating heads for specific syntactic relations or using probing classifiers. 
\citet{Marecek2018ExtractingST} propose heuristic ways of converting attention scores to syntactic trees, but do not quantitatively evaluate their approach.  
For coreference, \citet{Voita2018ContextAwareNM} show that the attention of a context-aware neural machine translation system captures anaphora, similar to our finding for BERT. 

Concurrently with our work \citet{voita2019analyzing} identify syntactic, positional, and rare-word-sensitive attention heads in machine translation models.
They also demonstrate that many attention heads can be pruned away without substantially hurting model performance.
Interestingly, the important attention heads that remain after pruning tend to be ones with identified behaviors. 
\citet{michel2019sixteen} similarly show that many of BERT's attention heads can be pruned.
Although our analysis in this paper only found interpretable behaviors in a subset of BERT's attention heads, these recent works suggest that there might not be much to explain for some attention heads because they have little effect on model perfomance.  

\citet{Jain2019AttentionIN} argue that attention often does not ``explain" model predictions.
They show that attention weights frequently do not correlate with other measures of feature importance. 
Furthermore, attention weights can often be substantially changed without altering model predictions. 
However, our motivation for looking at attention is different: rather than explaining model predictions, we are seeking to understand information learned by the models.
For example, if a particular attention head learns a syntactic relation, we consider that an important finding from an analysis perspective even if that head is not always used when making predictions for some downstream task. 

\section{Conclusion}
We have proposed a series of analysis methods for understanding the attention mechanisms of models and applied them to BERT.
While most recent work on model analysis for NLP has focused on probing vector representations or model outputs, we have shown that a substantial amount of linguistic knowledge can be found not only in the hidden states, but also in the attention maps.
We think probing attention maps complements these other model analysis techniques, and should be part of the toolkit used by researchers to understand what neural networks learn about language. 

\section*{Acknowledgements}
We thank the anonymous reviews for their thoughtful comments and suggestions. Kevin is supported by a Google PhD Fellowship.

\bibliography{bert_attention}
\bibliographystyle{acl_natbib}

\end{document}